\documentclass[letterpaper]{article} 
\usepackage{aaai2027}

\usepackage[hyphens]{url}  
\usepackage{graphicx} 
\usepackage{natbib}  
\usepackage{caption} 
\usepackage{algorithm}
\usepackage{algorithmic}
\usepackage{amsmath}
\usepackage{amsfonts}
\usepackage{amssymb}
\usepackage{booktabs}
\usepackage{multirow}
\usepackage{tabularx}
\usepackage{array}
\usepackage{float}
\usepackage{subcaption}
\usepackage[table]{xcolor}
\definecolor{lightpurple}{RGB}{235,230,250}
\newcolumntype{Y}{>{\centering\arraybackslash}X}
\usepackage{booktabs}
\usepackage{amssymb}
\usepackage{xcolor}
\usepackage{multirow}
\usepackage{newfloat}
\usepackage{listings}
\DeclareCaptionStyle{ruled}{labelfont=normalfont,labelsep=colon,strut=off} 
\floatstyle{ruled}
\newfloat{listing}{tb}{lst}{}
\floatname{listing}{Listing}

\usepackage{booktabs}

\title{Refine Then Fusion: Training-Free 3D Point Cloud Adaptation with Priority Refinement and Multi-Modal Knowledge Fusion}
\author{
    Hang Cheng\equalcontrib,
    Yan Chen\equalcontrib,
    Mingyu Fan\equalcontrib,
    Long Zeng\corresponding
}
\affiliations{
    \textsuperscript{\rm 1}Tsinghua Shenzhen International Graduate School, Tsinghua University\\
    chenghan24@mails.tsinghua.edu.cn, chenyan23@mails.tsinghua.edu.cn, my-fan25@mails.tsinghua.edu.cn, zenglong@sz.tsinghua.edu.cn
}

\begin{document}

\maketitle

\begin{abstract}
Recent pre-trained foundation models provide rich multi-modal priors for downstream 3D vision tasks. However, the effectiveness of these representations in few-shot scenarios is limited by two fundamental challenges: High-dimensional features often contain substantial channel redundancy and task-irrelevant noise, while the reliability of different modalities varies across samples. Consequently, direct aggregation of heterogeneous representations overlooks sample-dependent modality reliability and may obscure the discriminative cues essential. To address these limitations, we propose \textbf{Refine Then Fusion (RTF)}, a training-free framework for few-shot 3D recognition. RTF first identifies discriminative feature channels by jointly modeling inter-class similarity and intra-class stability, thereby decoupling domain-specific knowledge refinement from the cached representations of pre-trained models. It then introduces a reliability-aware fusion mechanism that estimates sample-wise modality reliability from the distribution shifts induced by feature refinement, enabling adaptive aggregation of multi-modal representations. Furthermore, RTF constructs a memory cache that integrates instance-level support features with class-level prototypes to infer query labels. Extensive experiments on five benchmarks demonstrate that RTF consistently outperforms single-modal baselines, partial-fusion variants, and existing lightweight adaptation methods, achieving state-of-the-art few-shot 3D recognition performance without gradient optimization, additional training data, auxiliary training, or parameter updates.

\end{abstract}


\section{Introduction}

\begin{figure}[t]
\centering
\includegraphics[width=0.99\columnwidth]{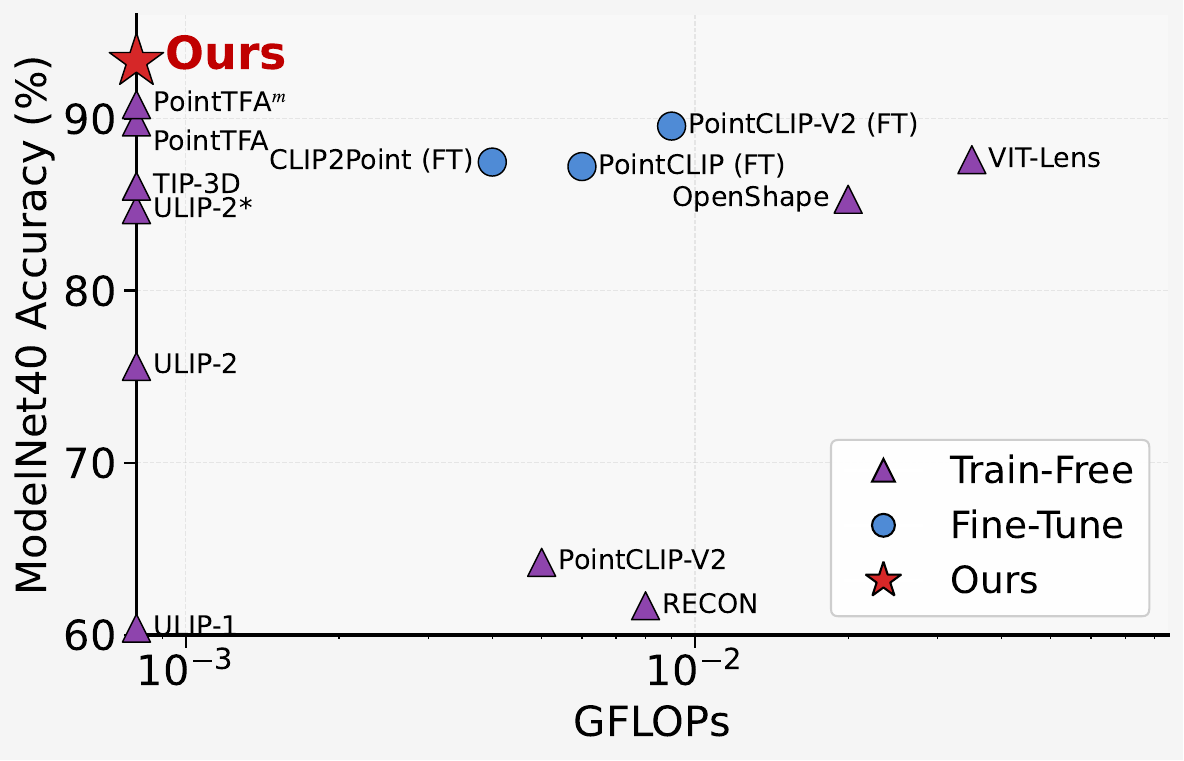} 
\caption{Comparison of accuracy and training GFLOPs on 16-shot ModelNet40 classification. Our method achieves competitive accuracy with substantially reduced optimization cost, demonstrating its effectiveness and efficiency for training-free adaptation.}
\label{fig1}
\end{figure}

\begin{figure}[t]
  \centering
  \includegraphics[width=\columnwidth]{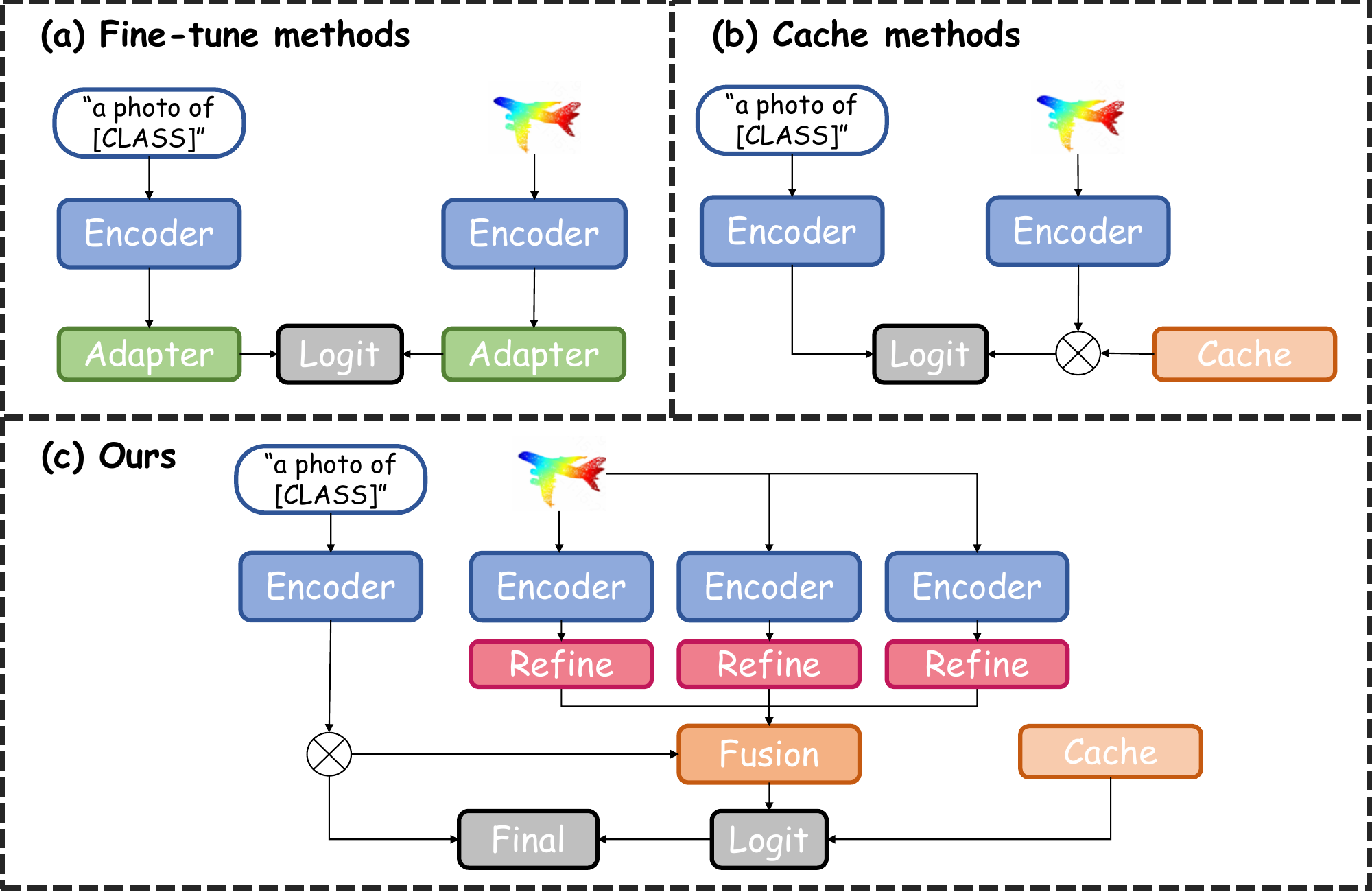}
  \caption{Comparison between previous adaptation paradigms and our RTF framework.}
  \label{fig:intro-method}
\end{figure}

Learning 3D object recognition with limited supervision remains a fundamental challenge in computer vision \cite{feng2025hyperbolic, khan2024enhancing}. Recently, multi-modal foundation models pre-trained on large-scale datasets \cite{huang2023clip2point, liu2024openshape}, such as ULIP \cite{xue2023ulip,xue2024ulip2} and PointCLIP \cite{zhang2022pointclip,zhu2023pointclipv2}, have demonstrated remarkable zero-shot capability in 3D understanding by providing rich transferable priors. However, adapting these representations to downstream few-shot scenarios remains challenging due to severe domain shifts and limited annotations. A straightforward solution is to fine-tune the foundation models or employ parameter-efficient adaptation strategies \cite{zhang2024taadapter, cheng2023metaadapter}. However, such optimization-based approaches inevitably require gradient updates, additional training procedures, and careful hyperparameter tuning, which are undesirable in data-scarce and resource-constrained scenarios. 

Pioneering efforts such as Tip-Adapter \cite{zhang2022tip} and its 3D extensions PointTFA \cite{wu_pointtfa}) have demonstrated the feasibility of training-free adaptation via key-value cache models. Nevertheless, we identify two fundamental bottlenecks in these existing architectures. \textbf{First, severe feature redundancy.} Although foundation models provide abundant multi-modal knowledge, such representations are not uniformly reliable for downstream tasks \cite{zhu2023not, lin2024mope}. First, high-dimensional features extracted from pre-trained models often contain substantial channel redundancy and task-irrelevant information, where only a subset of feature dimensions contribute to discriminative recognition. Directly retaining all feature channels not only introduces unnecessary computational overhead but may also dilute task-relevant cues stored in the representations. Second, different modalities exhibit varying reliability across samples due to their distinct semantic and geometric characteristics. 

\textbf{Second, naive multi-modal fusion.} Point clouds inherently benefit from diverse modalities---such as 3D semantics \cite{xue2023ulip}, 2D visual projections \cite{ji2024jm3dllm, wang2023jm3d}, and structural geometry \cite{zhang2023pointnn}. However, seamlessly integrating such heterogeneous modalities in few-shot scenarios poses a formidable challenge. Due to the inherent severe self-occlusion and high-dimensional sparsity of 3D data, forcibly coercing these representations into a unified embedding space via early feature fusion inevitably triggers the curse of dimensionality \cite{wu2026pointtfam}. More critically, it induces negative information gain—catastrophic noise from modalities operating within their cognitive blind spots can easily overwhelm the correct predictive signals from confident modalities. 

To address these challenges, we propose \textbf{Refine Then Fusion (RTF)}, a training-free framework that follows a simple yet effective principle: \textit{refine informative knowledge before adaptively fusing heterogeneous representations}. Specifically, RTF introduces a {Priority-guided Channel Refinement (PCR)} module, which identifies discriminative feature channels by jointly modeling inter-class similarity and inter-class variance. PCR selectively preserves task-relevant dimensions while suppressing redundant information in pre-trained representations, providing cleaner features for subsequent adaptation.

Building upon the refined representations, RTF develops a {Reliability-aware Multi-Modal Fusion (RAMF)} mechanism that estimates modality reliability according to refinement-induced distribution shifts. By measuring the consistency of each modality before and after refinement, RAMF adaptively aggregates complementary multi-modal representations and emphasizes stable knowledge while reducing the impact of unreliable modalities. Furthermore, RTF introduces a {Hybrid Memory Cache (HMC)} that integrates instance-level support retrieval with class-level prototype memory. By combining fine-grained sample matching and category-level abstraction, HMC improves query inference robustness under few-shot scenarios without additional parameter optimization.

In summary, the main contributions of our work are as follows: (1) We propose {Refine Then Fusion (RTF)}, a training-free framework that refines informative representations before adaptive multi-modal fusion for few-shot 3D recognition. (2) We introduce {Priority-guided Channel Refinement (PCR)}, which exploits inter-class similarity and variance to identify discriminative channels, suppressing redundant dimensions while preserving task-relevant knowledge. (3) We develop {Reliability-aware Multi-Modal Fusion (RAMF)} and {Hybrid Memory Cache (HMC)}, which respectively enable reliability-driven modality aggregation and complementary instance-prototype retrieval for robust query inference. (4) RTF achieves state-of-the-art performance among training-free methods on five benchmarks without any training costs.

\section{Related Work}
\subsection{3D Pre-trained Foundation Models}
Recent 3D point-cloud understanding increasingly relies on transferring vision--language priors from large-scale pretrained models, mitigating the need for training 3D encoders from scratch. PointCLIP-V1/V2 \cite{zhang2022pointclip, zhu2023pointclipv2} perform CLIP-based zero-shot recognition via multi-view depth projections and enhanced textual prompts, while CLIP2Point, ViT-Lens \cite{lei2024vitlens}, openshape\cite{liu2024openshape}and ULIP \cite{xue2023ulip, xue2024ulip2} and CG3D \cite{hegde2023cg3d} improve cross-modal adaptation by aligning 3D representations with pretrained image--text encoders. Despite different designs, these methods share the same premise: leveraging upstream multimodal knowledge for 3D representation learning. Building on such pretrained encoders, we study efficient downstream adaptation without retraining.

\subsection{Parameter-Efficient Fine-Tuning and Few-Shot Adaptation}
Parameter-Efficient Fine-Tuning (PEFT) has emerged as an effective paradigm for adapting foundation models to few-shot downstream tasks. In 2D vision-language models, methods such as CoOp~\cite{zhou2022coop}, CoCoOp~\cite{zhou2022cocoopl}, VPT~\cite{jia2022vpt}, and CLIP-Adapter~\cite{gao2021clipadapter} introduce learnable prompts or adapters, while prior-guided approaches including Tip-Adapter~\cite{zhang2022tipadapter}, Causal-FS~\cite{lin2024cause}, and Tip-X~\cite{udandarao2023tipX} exploit few-shot support sets through explicit key-value caches. Inspired by these advances, recent works extend few-shot adaptation to 3D perception, including CrossPoint~\cite{afham2022crosspoint}, PointTFA~\cite{wu_pointtfa, wu2026pointtfam}, GAPrompt~\cite{ai2025gaprompt}, TopAdapter~\cite{wang2026topadapter}, and UPP~\cite{ai2025upp}. However, existing 3D adaptation methods either rely on gradient-based optimization or suffer from insufficient feature refinement. In contrast, our RTF performs training-free feature refinement to suppress noisy and redundant information, enabling more effective adaptation.

\subsection{Multi-Modal Knowledge Fusion}
Multi-modal representation learning exploits complementary knowledge across heterogeneous modalities through cross-modal interaction. Attention-based fusion methods capture rich modality dependencies but introduce high computational costs~\cite{sun2023catnet, zhang2023transformergan, xing2023dualmodality, tan2019lxmert}. In contrast, CLIP-style dual encoders learn scalable shared embedding spaces via large-scale contrastive pre-training, enabling effective knowledge transfer for open-vocabulary tasks~\cite{ramesh2021dalle, zhou2024admnet}. Inspired by these advances, recent works extend multi-modal knowledge transfer to 3D perception. GPT4Point~\cite{qi2024gpt4point} and LL3DA~\cite{chen2024ll3da} incorporate language models for point cloud understanding, while others explore semantic interaction and visual-language transfer for 3D representations~\cite{xu2024multi, zhou2025crossmv, jiao2025clipgs}. However, most existing approaches still rely on additional training or adaptation, motivating our training-free framework.

\section{Proposed Method}

\begin{figure*}[t]
  \centering
  \includegraphics[
    width=\textwidth,
    trim=0mm 0mm 0mm 0mm,
    clip
  ]{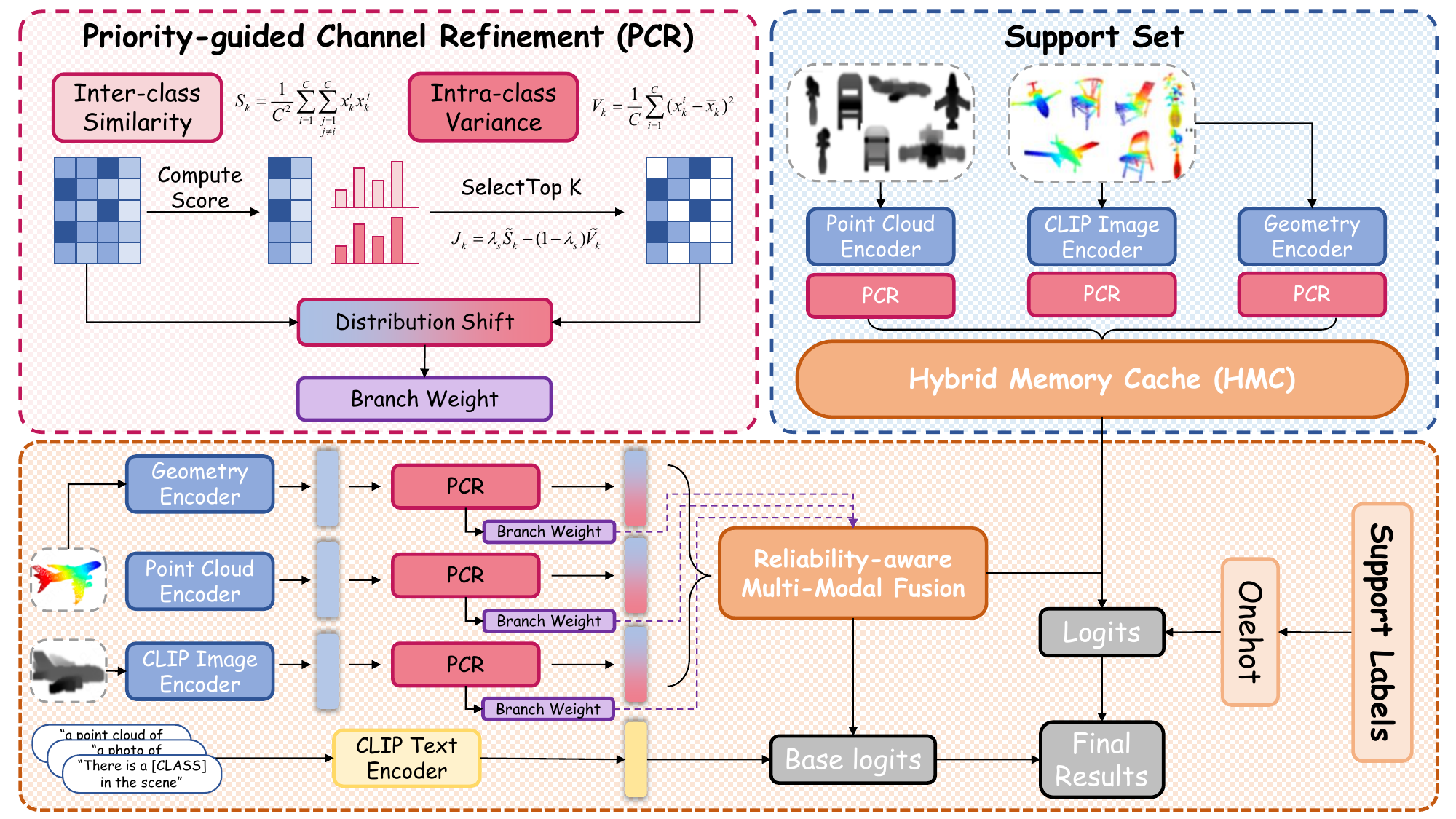}
  \caption{Overview of the proposed RTF framework, consisting of Priority-guided Channel Refinement, Reliability-aware Multi-Modal Fusion, and Hybrid Memory Cache for training-free 3D adaptation.}
  \label{fig:framework}
\end{figure*}

In this section, we present the proposed \textbf{Refine Then Fusion (RTF)} framework for training-free 3D point cloud adaptation. As illustrated in Fig.~\ref{fig:framework}, RTF integrates three complementary modalities and operates without gradient optimization or parameter updates. We first introduce the multi-modal representation extraction process. Then, we present the \textbf{Priority-guided Channel Refinement (PCR)} module, which identifies and preserves discriminative feature dimensions while suppressing task-irrelevant redundancy. Next, we introduce the \textbf{Reliability-aware Multi-Modal Fusion (RAMF)} mechanism, which adaptively aggregates heterogeneous representations according to modality reliability. Finally, we describe the \textbf{Hybrid Memory Cache (HMC)} strategy, which combines instance-level features and class-level prototypes for efficient query inference.

\subsection{Multi-Modal Representation}

Given an input 3D point cloud $\mathcal{P}$, we construct a comprehensive representation by leveraging three complementary modalities, denoted as $\mathcal{M} = \{ \text{3D}, \text{2D}, \text{Geo} \}$.

\begin{itemize}
    \item \textbf{3D Semantics:} We employ a pre-trained 3D foundation model (e.g., ULIP) to extract holistic semantic features $\mathbf{f}_\text{3D} \in \mathbb{R}^{D_\text{3D}}$.
    
    \item \textbf{2D Visual Projections:} To capture complementary visual cues, we project $\mathcal{P}$ into multi-view depth/color images and extract visual representations using a pre-trained Vision Transformer (ViT), yielding $\mathbf{f}_\text{2D} \in \mathbb{R}^{D_\text{2D}}$.
    
    \item \textbf{Local Geometry:} To explicitly encode structural information independent of pre-training biases, we employ a non-parametric Point-NN to extract geometric features $\mathbf{f}_\text{Geo} \in \mathbb{R}^{D_\text{Geo}}$.
\end{itemize}

All extracted features are $L_2$-normalized and projected onto the unit hypersphere.
\subsection{Priority-guided Channel Refinement}
\label{sec:refinement}

Pre-trained representations contain rich transferable knowledge, but many feature channels are irrelevant or unstable for specific few-shot tasks. To remove such redundancy without introducing learnable parameters, we propose the \textbf{Priority-guided Channel Refinement (PCR)} module, which selects discriminative channels according to their inter-class separability.

Given a support set with $N_s$ samples from $C$ categories, we first construct class prototypes $\mathbf{X} \in \mathbb{R}^{C \times D_m}$ for each modality $m$. For zero-shot aligned models such as ULIP, prototypes are obtained from encoded text prompts, while for visual and geometric modalities they are computed by averaging support features within each class.

For the $k$-th channel, PCR evaluates its discriminative priority by jointly considering inter-class similarity $S_k$ and inter-class variance $V_k$. A desirable channel should exhibit low similarity across different classes and high variance among class prototypes. Therefore, we define the priority score as:
\begin{equation}
    J_k = \lambda_s \tilde{S}_k - (1-\lambda_s)\tilde{V}_k
\end{equation}
where $\lambda_s \in [0,1]$ controls the trade-off between the two criteria, and $\tilde{(\cdot)}$ denotes Min-Max normalization for scale alignment. Specifically, the inter-class similarity $S_k$ measures the average pairwise correlation of the $k$-th channel across different class prototypes, which can be efficiently computed as:
\begin{equation}
    S_k=\frac{1}{C(C-1)}
    \left[
    \left(\sum_{i=1}^{C}X_{i,k}\right)^2
    -
    \sum_{i=1}^{C}X_{i,k}^{2}
    \right].
\end{equation}
The inter-class variance $V_k$ is calculated as the variance of the $k$-th channel over the $C$ class prototypes, reflecting its capability to distinguish different categories.

Channels with smaller $J_k$ values are considered more discriminative. PCR retains the top-$K$ channels with the lowest priority scores:
\begin{equation}
    \mathbf{f}'_m=\Phi_{\text{TopK}}(\mathbf{f}_m,J^m)\in\mathbb{R}^{K},
\end{equation}
where $K=\lfloor\rho\times D_m\rfloor$ is controlled by the keep ratio $\rho$. This parameter-free refinement suppresses task-irrelevant dimensions while preserving discriminative information for downstream recognition.

\subsection{Reliability-aware Multi-Modal Fusion}
\label{sec:fusion}

Different modalities exhibit varying reliability across samples due to their distinct semantic, visual, and geometric characteristics. Direct feature averaging may introduce unreliable information and weaken discriminative cues. To address this issue, we propose the \textbf{Reliability-aware Multi-Modal Fusion (RAMF)} mechanism, which estimates modality reliability according to refinement-induced distribution shifts.

\textbf{Key Observation.} A reliable modality should preserve consistent semantic affinity after removing redundant channels. If feature refinement substantially changes the similarity distribution between a query and the support memory, the original modality likely relies on unstable or task-irrelevant dimensions. Conversely, a stable distribution indicates robust discriminative information.

Given a query feature $\mathbf{f}^{q}_{m}$ and support memory $\mathbf{F}^{s}_{m}\in\mathbb{R}^{N_s\times D_m}$, the pre-refinement affinity distribution is computed as:
\begin{equation}
    \mathbf{p}^{m}_{pre}
    =
    \text{Softmax}
    \left(
    \mathbf{f}^{q}_{m}
    (\mathbf{F}^{s}_{m})^\top
    \right).
\end{equation}

After PCR, the refined affinity distribution $\mathbf{p}^{m}_{post}$ is obtained using $\mathbf{f}'^{q}_{m}$ and ${\mathbf{F}'}^{s}_{m}$. The reliability of modality $m$ is measured by the KL divergence between the two distributions:
\begin{equation}
    \mathcal{D}^{m}_{KL}
    =
    \sum_{i=1}^{N_s}
    \mathbf{p}^{m}_{pre}[i]
    \log
    \frac{
    \mathbf{p}^{m}_{pre}[i]
    }{
    \mathbf{p}^{m}_{post}[i]
    }.
\end{equation}

To obtain a stable modality-level reliability estimation under few-shot settings, we calculate the expected distribution shift:
\begin{equation}
    \bar{\mathcal{D}}^{m}_{KL}
    =
    \mathbb{E}_{q}
    [
    \mathcal{D}^{m}_{KL}
    ].
\end{equation}

The fusion weight is then assigned through a temperature-scaled softmax:
\begin{equation}
    w^{m}
    =
    \frac{
    \exp(-\tau\bar{\mathcal{D}}^{m}_{KL})
    }
    {
    \sum_{j\in\mathcal{M}}
    \exp(-\tau\bar{\mathcal{D}}^{j}_{KL})
    },
\end{equation}
where $\tau>0$ controls the weighting sharpness.

Finally, refined features with different dimensions are zero-padded into a unified semantic space of dimension $D_{max}=\max_j(D_j)$ and normalized. The fused representation is obtained as:
\begin{equation}
    \mathbf{f}_{fused}
    =
    \frac{
    \sum_{m\in\mathcal{M}}
    w^{m}\tilde{\mathbf{f}}'_m
    }
    {
    \left\|
    \sum_{m\in\mathcal{M}}
    w^{m}\tilde{\mathbf{f}}'_m
    \right\|_2
    },
\end{equation}
where $\tilde{\mathbf{f}}'_m$ denotes the padded refined representation.

\subsection{Hybrid Memory Cache}

After RAMF, RTF performs non-parametric prediction through the proposed \textbf{Hybrid Memory Cache (HMC)}. In few-shot scenarios, instance-level retrieval preserves fine-grained sample characteristics but may suffer from incomplete intra-class coverage due to limited support samples. To address this limitation, HMC combines instance memory with a class-level prototype memory, capturing both local sample similarity and global category variations.

Given fused support features $\mathcal{F}_s=\{(f_i,y_i)\}_{i=1}^{N_s}$, the instance memory directly stores support representations and performs similarity-based voting:
\begin{equation}
z_{ins}
=
\sum_i
\exp(-\gamma(1-q^\top f_i))y_i ,
\end{equation}
where $q$ denotes the query feature.

To enhance category-level robustness, HMC further constructs a multi-prototype memory by clustering features within each class on the hypersphere using spherical K-means:
\begin{equation}
\mathcal{M}_{cls}=\{(p_{c,k},c)\},
\end{equation}
where $p_{c,k}$ denotes the $k$-th prototype of class $c$. The prototype memory prediction is obtained through the same similarity-based voting:
\begin{equation}
z_{cls}
=
\sum_{c,k}
\exp(-\gamma(1-q^\top p_{c,k}))y_c .
\end{equation}

Since instance retrieval and prototype retrieval provide complementary but potentially uncertain predictions, HMC adaptively combines them according to prediction entropy:
\begin{equation}
\alpha=
\frac{\exp(-H(z_{cls}))}
{\exp(-H(z_{ins}))+\exp(-H(z_{cls}))},
\end{equation}
where lower entropy indicates higher prediction confidence. The final prediction is computed as:
\begin{equation}
z=z_{ins}+\alpha z_{cls}.
\end{equation}

HMC integrates fine-grained instance discrimination and class-level semantic abstraction without parameter optimization, providing robust inference under few-shot settings.

\section{Experiments}

\begin{table*}[t]
  \centering
  \small
  \label{tab:main}
  \centering
  \setlength{\tabcolsep}{2.5pt}
  \begin{tabular}{llccccccc}
    \toprule
    Method & Conditions (K-shot) & 2D-Modal & 3D-Modal & ModelNet40 & ModelNet10 & OBJ\_ONLY & OBJ\_BG & OBJ\_T50RS \\
    \midrule
    PointCLIP & \textbf{Train-Free} (0) & \checkmark & \checkmark & 20.18 & 30.23 & 19.28 & 21.34 & 15.38 \\
    CLIP2Point & \textbf{Train-Free} (0) & \checkmark & \checkmark & 49.38 & 66.63 & 30.46 & 35.46 & 23.32 \\
    PointCLIP-V2 & \textbf{Train-Free} (0) & \checkmark & \checkmark & 64.22 & 73.13 & 50.09 & 41.22 & 35.36 \\
    RECON & \textbf{Train-Free} (0) & \checkmark & \checkmark & 61.70 & 75.60 & 43.70 & 40.40 & 30.50 \\
    OpenShape & \textbf{Train-Free} (0) & \checkmark & \checkmark & 85.30 & - & - & 56.70 & - \\
    VIT-Lens & \textbf{Train-Free} (0) & \checkmark & \checkmark & 87.60 & - & - & 60.10 & - \\
    ULIP-1 & \textbf{Train-Free} (0) & - & \checkmark & 60.40 & - & - & 48.50 & - \\
    ULIP-2 & \textbf{Train-Free} (0) & - & \checkmark & 75.60 & - & - & - & - \\
    ULIP-2* & \textbf{Train-Free} (0) & - & \checkmark & 84.70 & - & - & - & - \\
    Point-NN & \textbf{Train-Free} (Full) & - & \checkmark & 81.80 & - & 71.10 & 74.90 & 64.90 \\
    Seg-NN & \textbf{Train-Free} (Full) & - & \checkmark & 84.20 & - & - & - & - \\
    TIP-3D & \textbf{Train-Free} (16) & - & \checkmark & 86.06 & 89.76 & 73.49 & 75.56 & 59.61 \\
    \midrule
    CLIP2Point & \textbf{Fine-Tune} (16) & \checkmark & \checkmark & 87.46 & - & - & - & - \\
    PointCLIP & \textbf{Fine-Tune} (16) & \checkmark & \checkmark & 87.20 & - & - & - & - \\
    PointCLIP-V2 & \textbf{Fine-Tune} (16) & \checkmark & \checkmark & 89.55 & - & - & - & - \\
    \midrule

    PointTFA & \textbf{Train-Free} (16) & - & \checkmark & 89.79 & 92.62 & 80.90 & 82.10 & 67.18 \\
    {PointTFA$^{m}$} & {Train-Free} (16) & \checkmark & \checkmark & {90.80} & \textbf{93.94} & {85.54} & {84.51} & {72.03} \\
    \rowcolor{lightpurple}
     \textbf{Ours} & \textbf{Train-Free} (16) & \checkmark & \checkmark & \textbf{91.05} & {92.84} & \textbf{87.26} & \textbf{84.85} & \textbf{73.94} \\
    \midrule

    PointTFA [33] & {Train-Free} (Full) & - & \checkmark & 90.88 & {93.17} & {83.48} & 84.85 & 68.22 \\
    {PointTFA$^{m}$} & {Train-Free} (Full) & \checkmark & \checkmark & {91.33} & 92.96 & 83.00 & {85.40} & {73.14} \\
    \rowcolor{lightpurple}
    \textbf{Ours} & \textbf{Train-Free} (Full) & \checkmark & \checkmark & \textbf{92.91} & \textbf{93.50} & \textbf{91.57} & \textbf{90.36} & \textbf{81.85} \\
    \bottomrule
  \end{tabular}
\caption{Comparison with state-of-the-art methods under few-shot and full-shot 3D recognition settings. ``2D-Modal'' indicates the use of image inputs during inference, while ``3D-Modal'' denotes the use of point clouds. ULIP-2$^{*}$ represents the model pre-trained on the large-scale Objaverse dataset. Best results are highlighted in bold.}
\label{tab:main}
\end{table*}

\subsection{Experiment Setup}
\subsubsection{Datasets}
We evaluate top-1 classification accuracy under training-free few-shot settings on three downstream benchmarks. ModelNet10 \cite{wu2014modelnet} contains 4,899 synthetic 3D models from 10 categories, with 3,991 training and 908 test samples. ModelNet40 \cite{wu2014modelnet} extends it to 40 categories and comprises 9,843 training and 2,468 test samples. ScanObjectNN \cite{uy2019scanobj} contains 2,902 real-world object instances from 50 categories and provides three increasingly challenging variants: OBJ\_ONLY, OBJ\_BG, and OBJ\_T50RS, which introduce background clutter and geometric perturbations to assess robustness under realistic scanning conditions.

\subsection{Implementation Details}
Following previous few-shot 3D recognition protocols, we evaluate our method under the $K$-shot setting, where $K \in \{1,2,4,8,16\}$, using the same support and query splits for all compared methods. We employ ULIP-2 \cite{xue2024ulip2} as the 3D semantic encoder, ViT \cite{dosovitskiy2020vit} as the 2D visual encoder, and PointNN \cite{zhang2023pointnn} as the geometric encoder. For Priority Channel Refinement, the channel keep ratio $\rho$ is set to 0.5 by default for all modalities, and the trade-off coefficient $\lambda_s$ between inter-class similarity and variance is set to 0.5. The KL-guided fusion temperature $\tau$ is empirically set to 1.0. All experiments are conducted using the same hyperparameter configuration across different datasets without dataset-specific tuning.

\subsection{Comparison to State-of-the-Art Methods}
We compare RTF with representative zero-shot, fine-tuning-based, and training-free 3D recognition methods, including PointCLIP~\cite{zhang2022pointclip}, CLIP2Point~\cite{huang2023clip2point}, RECON~\cite{qi2023recon}, OpenShape~\cite{liu2024openshape}, ViT-Lens~\cite{lei2024vitlens}, ULIP-1~\cite{xue2023ulip}, ULIP-2~\cite{xue2024ulip2}, Point-NN~\cite{zhang2023pointnn}, Seg-NN~\cite{zhu2024pointnn}, TIP-3D~\cite{zhang2022tip}, PointTFA~\cite{wu_pointtfa}, and PointTFA$^{m}$~\cite{wu2026pointtfam}. As reported in Tab.~\ref{tab:main}, these methods cover different adaptation paradigms, including image-assisted, point-cloud-based, and fine-tuning strategies.

RTF achieves state-of-the-art performance on most benchmarks while maintaining a fully training-free pipeline. Under the 16-shot setting, RTF reaches 91.05\% accuracy on ModelNet40, outperforming PointTFA and PointTFA$^{m}$ by 1.26\% and 0.25\%, respectively. On the challenging ScanObjectNN benchmarks, RTF further improves over PointTFA$^{m}$ by 1.91\% on OBJ\_T50RS under 16-shot and 8.71\% under full-shot settings, demonstrating the robustness of PCR and RAMF in refining and aggregating reliable multi-modal knowledge. Although slightly below the strongest competitor on ModelNet10, RTF achieves the best performance on the remaining evaluated settings, highlighting the effectiveness of parameter-free knowledge refinement and fusion without model optimization.

\section{Ablation Studies}
\label{sec:ablation}

\subsection{Impact of Support Data Availability}

To evaluate the data efficiency of RTF, we vary the \textbf{Proportion of Training Set} used to construct the Hybrid Memory Cache (HMC) from $10\%$ to $100\%$. As shown in Tab.~\ref{tab:ratio}, increasing the available support data consistently improves performance across different benchmarks, as a larger memory cache provides broader coverage of the underlying data distribution. Notably, RTF maintains strong recognition accuracy with only $10\%$ of the training set, demonstrating its effectiveness in exploiting limited support information under few-shot scenarios.

\begin{table}[htbp]
\centering
\small
\setlength{\tabcolsep}{3.5pt}
\begin{tabular}{lcccccc}
\toprule
Proportion & 10\% & 20\% & 40\% & 60\% & 80\% & 100\% \\
\midrule
ModelNet10 & 89.73 & 91.25 & 91.96 & 92.29 & 92.62 & \textbf{93.50} \\
ModelNet40 & 89.71 & 91.00 & 91.25 & 91.57 & 92.26 & \textbf{92.91} \\
OBJ\_ONLY  & 84.17 & 85.44 & 86.06 & 87.61 & 88.47 & \textbf{91.57} \\
OBJ\_BG    & 80.72 & 84.51 & 87.78 & 87.44 & 88.64 & \textbf{90.36} \\
OBJ\_T50RS & 76.65 & 78.31 & 79.98 & 81.40 & 80.36 & \textbf{81.85} \\
\bottomrule
\end{tabular}
\caption{Impact of Different Support Set Sizes}
\label{tab:ratio}
\end{table}

\subsection{Effectiveness of Core Components}
\begin{figure*}[t]
  \centering
  \includegraphics[
    width=\textwidth,
    trim=0mm 0mm 0mm 0mm,
    clip
  ]{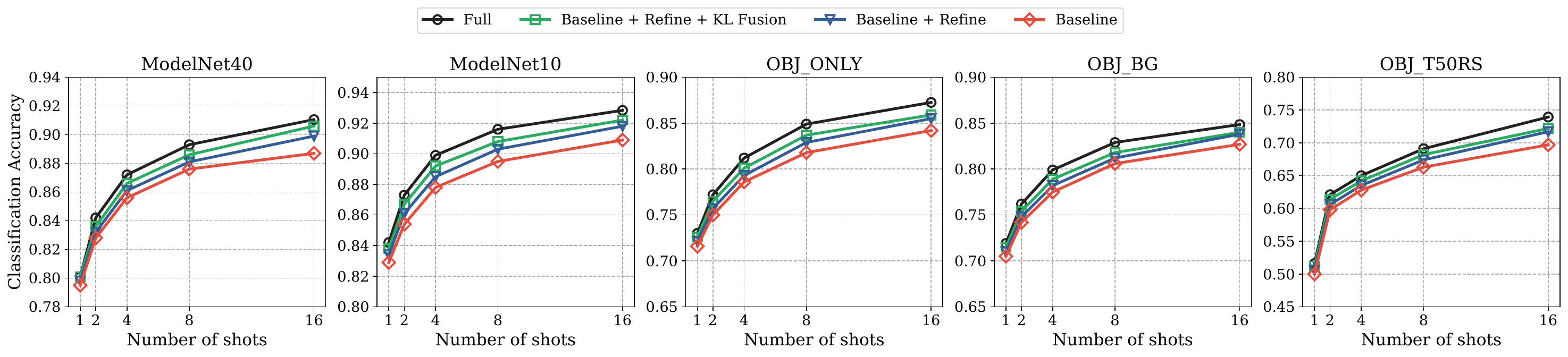}
\caption{Few-shot performance comparison of different core components ablation on multiple benchmarks.}
  \label{fig:components}
\end{figure*}

We conduct an ablation study to evaluate the contribution of each component in RTF. As shown in Fig.~\ref{fig:components}, performance consistently improves with the introduction of each component. Compared with direct averaging, {Priority-guided Channel Refinement (PCR)} brings the largest gain by removing task-irrelevant channels and obtaining cleaner representations. {Reliability-aware Multi-Modal Fusion (RAMF)} further improves performance by adaptively weighting modalities according to refinement-induced stability. Finally, {Hybrid Memory Cache (HMC)} provides additional gains by integrating instance-level retrieval and class-level prototype abstraction.

\subsection{Modality Contribution Analysis}
To analyze modality complementarity, we evaluate individual modalities and their combinations under the same setting. As shown in Tab.~\ref{tab:modality_ablation}, each modality captures distinct information with inherent limitations. The 3D semantic modality provides strong category understanding, while 2D visual and geometric modalities offer complementary appearance and structural cues. Combining modalities consistently improves performance; for example, adding geometry to 3D semantics increases OBJ\_T50RS accuracy from 77.97\% to 81.30\%. Finally, integrating all three modalities achieves the best results across all benchmarks, validating their complementary nature and motivating the adaptive fusion in RTF.

\begin{table*}[t]
\centering
\small

\renewcommand{\arraystretch}{1}
\begin{tabular}{ccc|ccccc}
\toprule
\multicolumn{3}{c|}{Modality}
& \multicolumn{5}{c}{Accuracy (\%)} \\
\cmidrule(lr){1-3}
\cmidrule(lr){4-8}
3D & 2D & Geometry & ModelNet40 & ModelNet10 & OBJ\_ONLY & OBJ\_BG & OBJ\_T50RS \\
\midrule
$\checkmark$ & $\times$ & $\times$
& 91.41 & 92.73 & 90.19 & 88.47 & 77.97 \\
$\times$ & $\checkmark$ & $\times$
& 83.14 & 87.56 & 70.05 & 74.18 & 54.75 \\
$\times$ & $\times$ & $\checkmark$
& 18.92 & 76.98 & 27.02 & 28.23 & 24.50 \\
$\checkmark$ & $\checkmark$ & $\times$
& 91.98 & 92.95 & 91.22 & 88.81 & 78.14 \\
$\checkmark$ & $\times$ & $\checkmark$
& 92.54 & 92.62 & 90.71 & 89.62 & 81.30 \\
$\times$ & $\checkmark$ & $\checkmark$
& 78.00 & 85.68 & 68.16 & 74.35 & 55.45 \\
$\checkmark$ & $\checkmark$ & $\checkmark$
& \textbf{92.91}
& \textbf{93.50}
& \textbf{91.57}
& \textbf{90.36}
& \textbf{81.85} \\

\bottomrule
\end{tabular}
\caption{Ablation study of different modality combinations.}
\label{tab:modality_ablation}
\end{table*}

\subsection{Comparison of Feature Fusion Strategies}
\begin{figure}[t]
\includegraphics[width=\columnwidth]{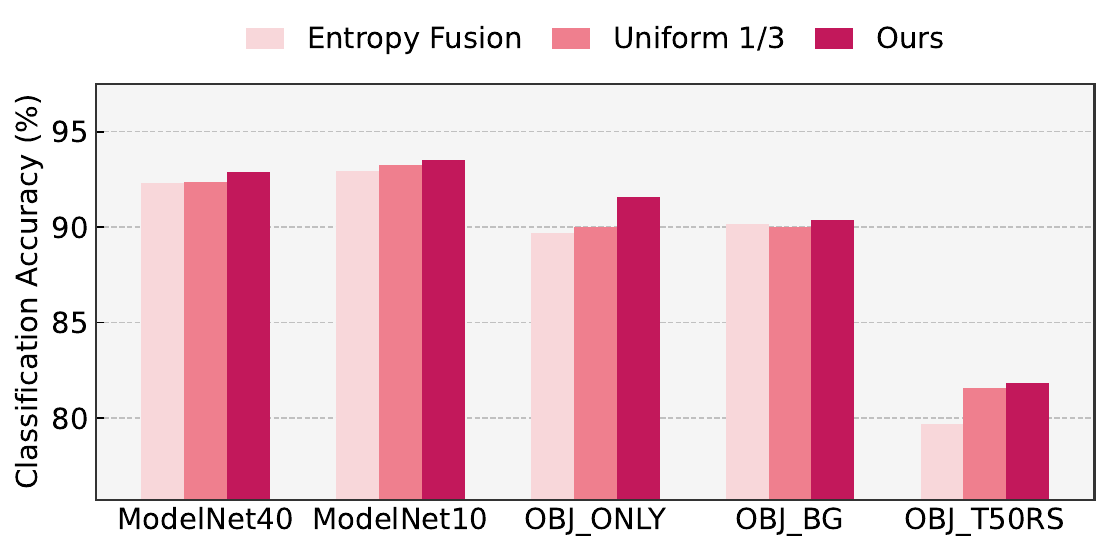} 
\caption{Comparison of different modality fusion strategies.}
\label{fig:fusion_strategy}
\end{figure}

To validate the effectiveness of the proposed reliability-aware weighting strategy, we compare KL-guided fusion with uniform averaging and entropy-based weighting. As shown in Fig.~\ref{fig:fusion_strategy}, entropy-based fusion provides limited improvements over uniform weighting, indicating that raw prediction confidence is insufficient for identifying reliable modalities. In contrast, KL-guided fusion consistently achieves the best performance by evaluating modality stability after PCR, demonstrating that refinement-induced distribution shifts provide a more reliable criterion for adaptive multi-modal aggregation.

\subsection{Comparison of Cluster Strategies}
\begin{figure}[t]
\includegraphics[width=\columnwidth]{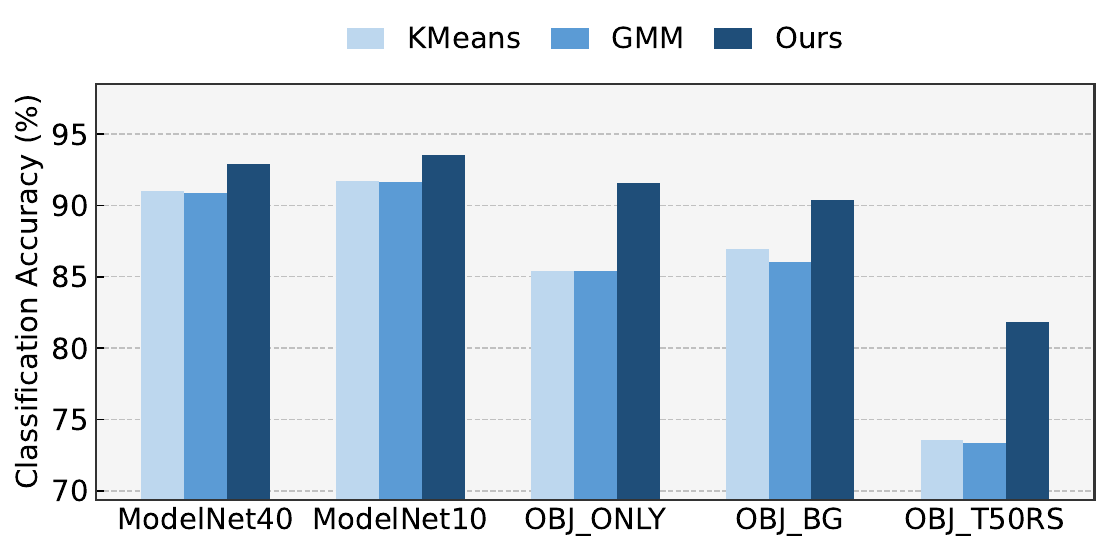} 
\caption{Comparison of different clustering strategies for class-level memory construction.}
\label{fig:cluster_strategy}
\end{figure}

To investigate the effectiveness of prototype construction in HMC, we compare different clustering strategies for building class-level memory. As shown in Fig.~\ref{fig:cluster_strategy}, Spherical K-Means consistently outperforms conventional K-Means and random prototype selection across all benchmarks. This improvement stems from the normalized feature space of pre-trained representations, where semantic similarity is better characterized by angular distance than Euclidean distance. By optimizing cosine similarity, Spherical K-Means produces more discriminative prototypes and better captures intra-class variations under few-shot settings.

\subsection{Sensitivity of Hyper-parameters}

\begin{figure}[t]
\includegraphics[width=\columnwidth]{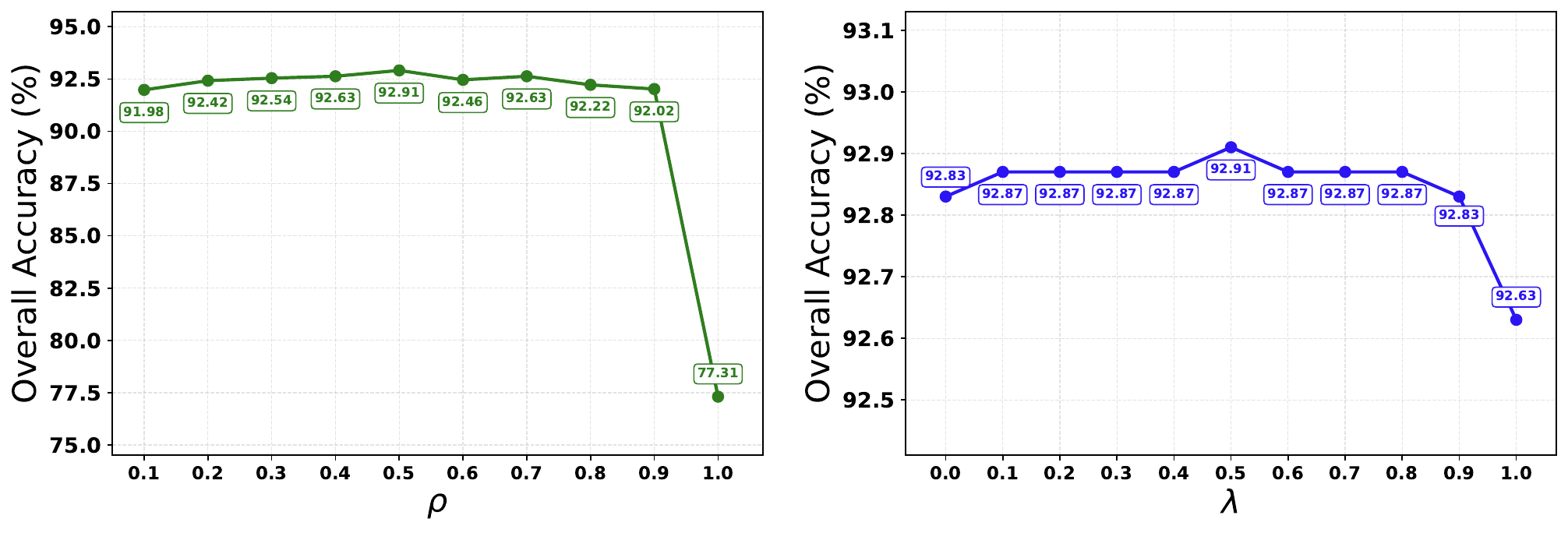} 
\caption{Sensitivity analysis of the channel keep ratio $\rho$ (left) and similarity-variance trade-off coefficient $\lambda_s$ (right).}
\label{fig:hyper_params}
\end{figure}

We investigate the sensitivity of two key hyper-parameters in PCR, including the channel keep ratio $\rho$ and the similarity-variance trade-off coefficient $\lambda_s$. As shown in Fig.~\ref{fig:hyper_params}, RTF maintains stable performance across a wide range of settings. The accuracy remains consistent when $\rho$ varies from 0.1 to 0.9, while overly retaining all channels ($\rho=1.0$) causes a clear degradation, confirming the necessity of removing redundant dimensions. Similarly, $\lambda_s$ exhibits limited influence within a broad range, indicating that PCR is robust to the balance between inter-class similarity and variance. These results demonstrate that the proposed refinement strategy does not rely on delicate parameter tuning.

\subsection{Robustness Against Noise}

To evaluate robustness under point cloud corruption, we inject Gaussian noise with $\sigma=0.01$ into point coordinates and compare different methods across benchmarks. As shown in Tab.~\ref{tab:noise_robustness}, noise severely degrades single-modality models, with ULIP-2 dropping from 75.6\% to 43.1\% on ModelNet40. In contrast, RTF consistently achieves the best performance under noisy inputs, outperforming PointTFA$^{m}$ across all datasets. This robustness is attributed to the complementary modalities, where the geometric branch provides stable structural cues and RAMF suppresses unreliable modalities according to refinement-induced reliability.

\begin{table}[t]
\centering

\resizebox{\linewidth}{!}{
\begin{tabular}{c|ccccc}
\toprule
\multirow{2}{*}{Dataset} & \multicolumn{1}{c}{ULIP-2} & \multicolumn{1}{c}{ULIP-2} & \multicolumn{1}{c}{PointTFA} & \multicolumn{1}{c}{PointTFA$^{m}$} & \multicolumn{1}{c}{Ours} \\ 
& (Noise-Free) & (+Noise) & (+Noise) & (+Noise) & (+Noise) \\
\midrule
ModelNet40 & 75.6 & 43.1 & 53.1 & 68.2 & \textbf{69.3} \\
ModelNet10 & 84.8 & 75.9 & 79.6 & 90.3 & \textbf{90.8} \\
OBJ\_ONLY & 53.4 & 31.5 & 44.0 & 47.0 & \textbf{49.5} \\
OBJ\_BG & 48.6 & 31.0 & 47.3 & 49.9 & \textbf{51.7} \\
OBJ\_T50RS & 40.4 & 22.7 & 27.2 & 32.1 & \textbf{35.6} \\
\bottomrule
\end{tabular}
}
\caption{Performance comparison under noise ($\sigma=0.01$) on different datasets.}
\label{tab:noise_robustness}
\end{table}

\section{Conclusion}

In this paper, we present {Refine Then Fusion (RTF)}, a training-free framework for few-shot 3D recognition that extracts reliable knowledge from pre-trained representations through a refine-then-fuse paradigm. Specifically, {Priority-guided Channel Refinement (PCR)} identifies discriminative dimensions by modeling inter-class similarity and variance, removing redundant information before adaptation. {Reliability-aware Multi-Modal Fusion (RAMF)} estimates modality reliability through refinement-induced distribution shifts and adaptively aggregates complementary semantic, visual, and geometric knowledge. {Hybrid Memory Cache (HMC)} further integrates instance-level retrieval and class-level prototypes for robust query inference. Extensive experiments demonstrate that RTF achieves state-of-the-art performance among training-free approaches without gradient optimization or parameter updates, highlighting the effectiveness of non-parametric knowledge refinement and fusion for efficient 3D adaptation.

\bibliography{aaai2027}


\end{document}